\documentclass[11pt,a4paper]{article}
\usepackage[hyperref]{An_Algorithm_for_Routing_Capsules_in_All_Domains}
\usepackage{times}
\usepackage{latexsym}
\usepackage{graphicx}
\usepackage{caption}
\usepackage{amsmath}
\usepackage{amsfonts}
\usepackage{booktabs}
\usepackage[bottom]{footmisc}
\usepackage{url}
\usepackage[linesnumbered,lined,algoruled]{algorithm2e}

\graphicspath{{figures/}}

\aclfinalcopy 

\title{An Algorithm for Routing Capsules in All Domains \\ with Sample Applications in Vision and Language}

\author{Franz A. Heinsen \\
  {\tt franz@glassroom.com} \\ }

\date{September [30], 2019}

\newcommand{\suptag}[1]{^{{\scriptscriptstyle\tt (#1)}}}
\newcommand{\cov}{\suptag{cov}}  
\newcommand{\inp}{\suptag{inp}}  
\newcommand{\out}{\suptag{out}}  

\newcommand{\showeqforV}[1]{
	\begin{cases}
		\sum_d W_{ijdh} \, \mu\inp_{icd} + B_{ijch},
		#1 \text{\small if $n\inp$ fixed}
		\\
		\sum_d W_{jdh} \, \mu\inp_{icd} + B_{jch},
		#1 \text{\small otherwise}
	\end{cases}
}

\newcommand{\showeqfora}[1]{
	\begin{cases}
		\sum_i \beta\suptag{use}_{ij} D\suptag{use}_{ij} - 
		\sum_i \beta\suptag{ign}_{ij} D\suptag{ign}_{ij},
		#1 \text{\small if $n\inp$ fixed} \\
		
		\sum_i \beta\suptag{use}_j D\suptag{use}_{ij} -
		\sum_i \beta\suptag{ign}_j D\suptag{ign}_{ij},
		#1 \text{\small otherwise}
	\end{cases}
}

\begin{document}
\maketitle

\begin{abstract} 
Building on recent work on capsule networks, we propose a new, general-purpose form of ``routing by agreement'' that activates output capsules in a layer as a function of their net benefit to use and net cost to ignore input capsules from earlier layers. To illustrate the usefulness of our routing algorithm, we present two capsule networks that apply it in different domains: vision and language.\footnote{
	In both domains, we use the same routing code, available at \href{https://github.com/glassroom/heinsen\_routing}{https://github.com/glassroom/heinsen\_routing} along with pretrained models and replication instructions.
} The first network achieves new state-of-the-art accuracy of \textbf{99.1\%} on the smallNORB visual recognition task with fewer parameters and an order of magnitude less training than previous capsule models, and we find evidence that it learns to perform a form of ``reverse graphics.'' The second network achieves new state-of-the-art accuracies on the root sentences of the Stanford Sentiment Treebank: {\bf 58.5\%} on fine-grained and {\bf 95.6\%} on binary labels with a single-task model that routes frozen embeddings from a pretrained transformer as capsules. In both domains, we train with the same regime.
\end{abstract}

\section{Introduction}\label{sec:introduction}

Capsule networks with routing by agreement can be more effective than convolutional neural networks for segmenting highly overlapping images \cite{DBLP:journals/corr/abs-1710-09829} and for generalizing to different poses of objects embedded in images and resisting white-box adversarial image attacks \cite{46653}, typically requiring fewer parameters but more training and computation.

A capsule is a group of neurons whose outputs represent different properties of the same entity in different contexts. Routing by agreement is an iterative form of clustering in which a capsule detects an entity by looking for agreement among votes from input capsules that have already detected parts of the entity in a previous layer.

\subsection*{Proposed Routing Algorithm}

Here, we propose a new, general-purpose form of ``EM routing'' \cite{46653} that uses the expectation-maximization (EM) algorithm to cluster similar votes from input capsules to output capsules. Each output capsule iteratively maximizes the probability of input votes assigned to it, given its probabilistic model.

Our EM routing algorithm has multiple differences compared to previous ones. The most significant difference is that we compute each output capsule's activation by applying a logistic function to the difference between a \emph{net benefit to use} and a \emph{net cost to ignore} ({\em i.e.}, not use) each input capsule. We compute the share of each input capsule used or ignored by each output capsule in a new procedure we call the \emph{D-Step}, executed between the E-Step and M-Step of each EM iteration. We are motivated by the intuitive notion that

\begin{quote}
\emph{``output capsules should benefit from the input data they use, and lose benefits from any input data they ignore,''}
\end{quote}

as they maximize the probability of votes from those input capsules they use in each EM iteration.

We simultaneously (a) optimize the entire layer for a training objective, (b) iteratively maximize the probability of input capsule votes each output capsule uses, and (c) iteratively maximize the probability of net input capsule benefits less costs in service of (a) and (b). We like to think of this mechanism as finding ``the combination of net benefits and costs that produces greater profit,'' or, more colorfully, maximizing \emph{``bang per bit.''}

Another significant difference of our routing algorithm, compared to previous ones, is that it accepts variable-size inputs, such as sequences of contextualized token embeddings in natural language applications. A contextualized token embedding is a special case of a capsule, one whose neuron outputs represent different properties of the same token id in different contexts.

\begin{figure}[t]
	\vskip 0.1in
	\begin{center}
		\centerline{\includegraphics{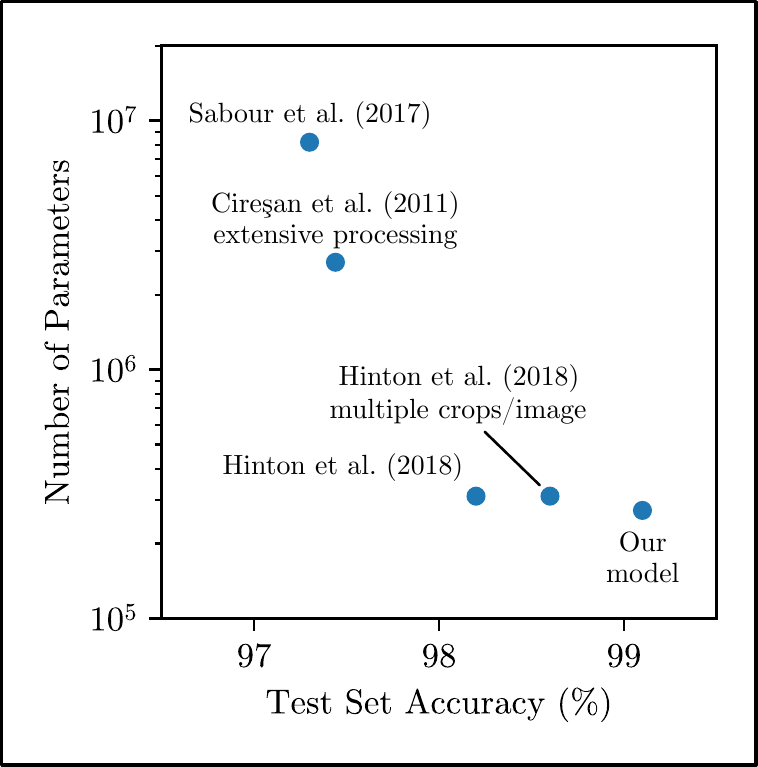}}
		\caption{Test set accuracy and number of parameters of models that have achieved state-of-the-art results on smallNORB visual recognition.}
		\label{fig:smallnorb_accuracy_vs_parameters}
	\end{center}
	\vskip -0.2in
\end{figure} 

\begin{figure}[t]
	\vskip 0.1in
	\begin{center}
		\centerline{\includegraphics{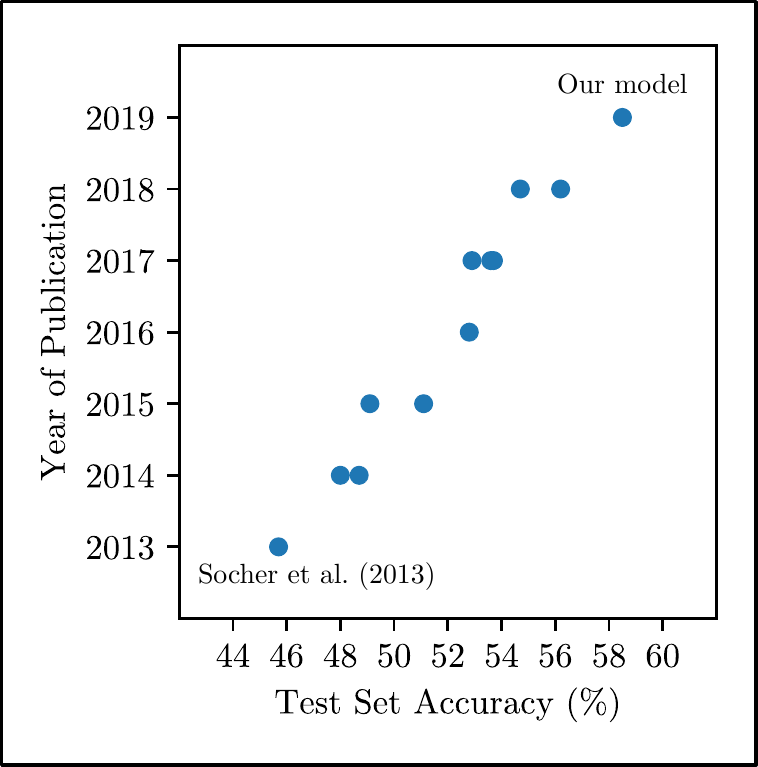}}
		\caption{Test set accuracy and publication year of models that have achieved state-of-the-art results on SST-5/R root sentence classification.}
		\label{fig:SST5R_accuracy_by_year}
	\end{center}
	\vskip -0.2in
\end{figure} 

\subsection*{Sample Applications in Two Domains}

For comparison with prior work on capsule networks, we evaluate our routing algorithm on the smallNORB visual recognition task \cite{LeCun:2004:LMG:1896300.1896315}, in which objects must be recognized from stereo images with varying azimuths, elevations, and lighting conditions. We construct a capsule network that routes pixel convolutions as capsules to achieve new state-of-the-art accuracy of 99.1\% on this task. Compared to the previous state of the art \cite{46653}, our smallNORB model has approximately 272,000 instead of 310,000 parameters, trains in 50 instead of 300 epochs, and accepts as input non-downsampled pairs of 96$\times$96 images, instead of 32$\times$32 downsampled crops that are nine times smaller. Also, we do not average over multiple crops per image to compute test accuracy. Fig. \ref{fig:smallnorb_accuracy_vs_parameters} shows how our model compares to prior state-of-the-art models on two criteria.

We find evidence that our smallNORB model learns to encode pose solely from pixel data and classification labels, {\em i.e.}, it learns its own form of ``reverse graphics'' without us explicitly having to optimize for it. See the visualization in Fig. \ref{fig:smallnorb_elevation_trajectories} (p. \pageref{fig:smallnorb_elevation_trajectories}) and the 24 plots and corresponding captions in Supp. Fig. \ref{fig:smallnorb_change_in_pose_vectors_azimuth} (p. \pageref{fig:smallnorb_change_in_pose_vectors_azimuth}) and Supp. Fig. \ref{fig:smallnorb_change_in_pose_vectors_elevation} (p. \pageref{fig:smallnorb_change_in_pose_vectors_elevation}).

For illustration of the general-purpose nature of our routing algorithm, we also evaluate it on a natural language task: classifying the root sentences of the Stanford Sentiment Treebank \cite{brusilovsky:socher2013recursive} into fine-grained (SST-5/R) and binary (SST-2/R) labels. (We add the ``/R'' designation to distinguish these root-sentence tasks from classification of phrases in the parse trees, because we have seen research that does not.)

For evaluation of our algorithm on SST-5/R and SST-2/R, we construct a capsule network with only approximately 140,000 parameters that routes frozen token embeddings from a pretrained transformer \cite{DBLP:journals/corr/VaswaniSPUJGKP17} as capsules. In our implementation, we use a GPT-2$_\text{large}$ \cite{GPT-2-Radford-et-al} as the pretrained transformer, the largest such model publicly available at the time of writing. Our SST model achieves new state-of-the-art test set accuracy of 58.5\% on SST-5/R, and new state-of-the-art test set accuracy for single-task models of 95.6\% on SST-2/R. Fig. \ref{fig:SST5R_accuracy_by_year} shows how our SST model compares to prior state-of-the-art models on SST-5/R.

\subsection*{Motivation}

As we show here, {\em we can achieve state-of-the-art results in more than one domain by stacking one or more layers of our routing algorithm atop, or between, blocks of non-iterative layers} (e.g., convolutional, self-attention, LSTM). Our motivation is to develop universal, composable learning algorithms that can adapt to any application.

\section{Notation}\label{sec:notation}

We use tensor notation, with all operations performed element-wise (Hadamard), implicitly assuming conventional broadcasting for any missing indexes. This notation is both more succint than linear algebra's and more natural to implement with frameworks like PyTorch and TensorFlow. For extra clarity, we do not use implicit (Einstein) summations nor raised (contravariant) indexes. See Tab. \ref{tab:notation_examples} for examples of our notation and their implementation in PyTorch.

\begin{table}[h]
	\small
	\begin{center}
		\begin{tabular}{@{}ll@{}}
			\toprule
			\bf Example & \bf Implementation in PyTorch \\
			\midrule
			\addlinespace[0.6em]
			$A_{ij} + B_{ijk}$ & {\tt A.unsqueeze(-1) + B} \\
			\addlinespace[0.6em]
			$A_{ij} B_{ijk}$ & {\tt A.unsqueeze(-1) * B}\quad or \\
			 & {\tt einsum("ij,ijk->ijk", A, B) } \\
			\addlinespace[0.6em]
			$\sum_i A_{ij} B_{ijk}$ & {\tt einsum("ij,ijk->jk", A, B)} \\
			\addlinespace[0.6em]
			$\sum_{ij} A_{ij} B_{ijk}$ & {\tt einsum("ij,ijk->k",\,A,\,B)} \\
			\addlinespace[0.6em]
			\bottomrule
		\end{tabular}
	\end{center}
	\caption{\label{tab:notation_examples}Examples of our notation and their implementation in PyTorch. In all examples, {\tt A} has shape $d_1 \times d_2$ and {\tt B} has shape $d_1 \times d_2 \times d_3$.}
\end{table}

\section{Our Routing Algorithm}\label{sec:our_routing_algorithm}

For brevity, we assume familiarity with both capsule networks and the matrix EM routing algorithm proposed by \citet{46653}, so we focus our discussion here mainly on those aspects of our work that are new. Also, while our algorithm generalizes to any probabilistic model that can be used in an expectation-maximization loop, we restrict our discussion here only to one case: a multidimensional Gaussian model.

As shown in Alg. \ref{alg:Our_EM_Routing_algorithm}, for a given sample, our algorithm dynamically routes $n\inp$ input capsules to $n\out$ output capsules, where $n\inp$ is specified in advance if samples are of fixed size or left unspecified if samples are of variable size. Input capsules are tensors of size $d\cov \times d\inp$, and output capsules are tensors of size $d\cov \times d\out$.

\begin{algorithm*}
	\caption{\textbf{Our Routing Algorithm}. Per sample, we route $n\inp$ input capsules of shape $d\cov \times d\inp$ to $n\out$ output capsules of shape $d\cov \times d\out$, where $n\inp$ is the number of detectable entity parts and $n\out$ is the number of detectable entities. All tensor operations shown are element-wise ({\em i.e.}, there are no Einstein summations) and implicitly assume conventional broadcasting for missing indexes. $f$ is the logistic function. Superscript text in parenthesis, such as $\inp$ and $\out$, denotes labels used for disambiguation. Tensor indexes are $i = (1, 2, \dots, n\inp)$, $j = (1, 2, \dots, n\out)$, $c = (1, 2, \dots, d\cov)$, $d = (1, 2, \dots, d\inp)$, $h = (1, 2, \dots, d\out)$.}
	\label{alg:Our_EM_Routing_algorithm}
	\KwIn{$( a\inp_i, \; \mu\inp_{icd} ).$}
	\KwOut{$( a\out_j, \; \mu\out_{jch}, \; (\sigma\out_{jch})^2 ).$}
	\BlankLine
	$V_{ijch} \longleftarrow \showeqforV{&}$\;
	\For{$n^{\scriptscriptstyle\tt (iters)}$ iterations}{
		\Begin(E-Step){
			\eIf{on first iteration}{
				$R_{ij} \longleftarrow \frac{1}{n\out}$\;
			}{
				$P_{ij} \longleftarrow
					\frac{1}{\sqrt{\prod_{ch} 2 \pi \left(\sigma\out_{jch}\right)^2 }}
					\exp \left( {-\sum_{ch}
						\frac{\left(V_{ijch} -\mu\out_{jch}\right)^2} 	{2\left(\sigma\out_{jch}\right)^2} }
					\right)$\;
				$R_{ij} \longleftarrow
					\frac{f\left(a\out_j\right) P_{ij}}
					{\sum_j f\left(a\out_j\right) P_{ij}}$\;
			}
		}
		\Begin(D-Step){
			$D\suptag{use}_{ij} \longleftarrow f(a\inp_i) R_{ij}$\;
			$D\suptag{ign}_{ij} \longleftarrow f(a\inp_i) - D\suptag{use}_{ij}$\;
		}
		\Begin(M-Step){
			$a\out_j \longleftarrow \showeqfora{&}$\;
			$\mu\out_{jch} \longleftarrow
				\frac{\sum_i D\suptag{use}_{ij} V_{ijch}}{\sum_i D\suptag{use}_{ij}}$\;	
			$(\sigma\out_{jch})^2 \longleftarrow
				\frac{\sum_i D\suptag{use}_{ij} \left( V_{ijch} - \mu\out_{jch} \right) ^2}{\sum_i D\suptag{use}_{ij}}$\;
		}
	}
\end{algorithm*}

Per sample, we accept as input two tensors: \emph{scores} and \emph{capsules} $(a\inp_i, \, \mu\inp_{icd})$. We return as output three tensors: \emph{scores}, \emph{capsules}, and \emph{variances} $(a\out_j, \, \mu\out_{jch}, \, (\sigma\out_{jch})^2)$. The indexes are

\begin{equation*}
\begin{aligned}
i & = (1, 2, \dots, n\inp), \\
j & = (1, 2, \dots, n\out), \\
c & = (1, 2, \dots, d\cov), \\
d & = (1, 2, \dots, d\inp), \\
h & = (1, 2, \dots, d\out).
\end{aligned}
\end{equation*}

Intuitively, we can think of $n\inp$ as the number of detectable parts, $n\out$ as the number of detectable entities, each consisting of or associated with one or more parts, $d\cov$ as the dimension of the covector space, or dual, of the spaces in which parts and entities have properties, and $d\inp$ and $d\out$ as the dimensions of part and entity properties, respectively, in those spaces.

For example, if we wanted to detect, say, dogs and cats embedded in images, from 64 detectable animal parts, the values of $n\inp$, $n\out$, $d\cov$, $d\inp$, and $d\out$ would be 64, 2, 4, 4, and 4, respectively. Each of the 64 input capsules and 2 output capsules would be a 4 $\times$ 4 matrix capable of representing the spatial relationship between the viewer of an image and objects embedded in the image.

In other domains, the dimensions $d\cov$ of the dual space, $d\inp$ of part properties, and $d\out$ of entity properties may be very different.

When $n\inp$ is unspecified, the number of detectable parts is variable. In that case, it might be desirable for input capsules themselves to have properties that represent their type and/or position. This is commonly done, for example, in language models, which add relative or absolute position information to token embeddings.

\subsection{Votes}\label{ssec:Computation-of-Votes}

Before starting the routing loop, we compute votes from each input to each output capsule,

\begin{equation}\label{eq:equation_for_V}
V_{ijch} = \showeqforV{&}
\end{equation}

where $V_{ijch}$ is a tensor of votes computed from the $i$th input capsule to the $j$th output capsule for each component $ch$ of the output capsule, and $W_{ijdh}$ and $B_{ijch}$ (or $W_{jdh}$ and $B_{jch}$, if $n\inp$ is unspecified) are parameters. We perform tensor contraction on index $d$ and compute element-wise operations, as indicated, along indexes $i$, $j$, $c$, and $h$, with conventional broadcasting implicitly assumed for any missing dimensions. For each input capsule $i$, we obtain a different $jch$ slice of votes for the $j$th output capsule, breaking symmetry.

Intuitively, the computation of $V_{ijch}$ in (\ref{eq:equation_for_V}) can be understood as $n\inp \times n\out$ simultaneous matrix-matrix multiplications, each applying a $d\out \times d\inp$ linear transformation to a $d\inp \times d\cov$ transposed input capsule, followed by addition of biases and then another transposition, to obtain $n\inp \times n\out$ votes of size $d\cov \times d\out$.

When $n\inp$ is left unspecified, we remove index $i$ from the parameters used to compute $V_{ijch}$, so we reduce their size by a factor of $n\inp$. In this case, the computation of $V_{ijch}$ applies $n\out$ simultaneous matrix-matrix multiplications to each input capsule, followed by addition of $n\out$ corresponding $ch$ biases.

\subsubsection{Adapting to Variable-Size Outputs}
\label{sssec:handling_variable_size_outputs}

A trivial adaptation of our algorithm, which we do not show, is to allow both $n\inp$ and $n\out$ to be unspecified, resulting in a variable number of input capsules voting for an equal number of output capsules. We can do this by removing indexes $ij$ from $W_{ijdh}$, reducing its size by a factor of $n\inp \times n\out$. In that case, the computation of $V_{ijch}$ would apply the same linear transformation to all input capsules, and would have to break symmetry via other means (e.g., by adding different biases).

\subsection{Routing Loop}\label{ssec:routing_iterations}

Unlike previous versions of EM routing, which on each iteration compute first an M-Step and then an E-Step procedure, our algorithm computes these procedures in the conventional order: first the E-Step and then the M-Step. This ordering removes the final, superfluous E-Step from the loop, and also, we believe, simplifies exposition.

We also introduce a new procedure, which we call the \emph{D-Step}, between the E-Step and M-Step. The D-Step computes the share of each input capsule's data used or ignored by each output capsule, for subsequent use in the computation of output scores $a\out_j$. These computations, described in \ref{ssec:D_Step} and \ref{ssec:M_Step}, represent our most significant departure from previous forms of EM routing.

Another difference is that in our algorithm, $a\inp_i$ and $a\out_j$ are ``pre-activation'' scores in the interval $[-\infty, \infty]$ on which we apply logistic functions as needed, at the last minute as it were, to compute activations. This trivial modification facilitates more flexible use of the ``raw'' values of $a\inp_i$ and $a\out_j$ by subsequent layers and/or objective functions, with more numerical stability. For example, a subsequent layer can apply a Softmax function to the output scores $a\out_j$ to induce a distribution over output capsule activations.

In the following subsections, we describe the computations performed by the E-Step, D-Step, and M-Step on each iteration, in order of execution, emphasizing those computations which are new or different with respect to previous work.

\subsection{E-Step}\label{ssec:E_Step}

At the start of each iteration, for each sample, our E-Step computes an $n\inp \times n\out$ tensor $R_{ij}$ of routing probabilities for assigning each input capsule $i$ to each output capsule $j$,

\begin{equation}\label{eq:equation_for_R}
R_{ij} =
\begin{cases}
  \frac{1}{n\out},
  & \text{\small on first iteration} \\
  \\
  \frac{f\left(a\out_j\right) P_{ij}}
  {\sum_j f\left(a\out_j\right) P_{ij}},
  & \text{\small otherwise}
\end{cases}
\end{equation}

where $f$ is the logistic function, $a\out_j$ is the $j$th output score computed in the previous iteration's M-Step, and $P_{ij}$ stands for

\begin{equation}
\prod_{ch} P \left( V_{ijch} \; \middle| \; V\out_{jch} \sim \mathcal{N} \left(\mu\out_{jch}, (\sigma\out_{jch})^2 \right) \right),
\end{equation}

the products of the probability densities of input capsule $i$'s votes for output capsule $j$'s $ch$ components, given output capsule $j$'s Gaussian model, updated in the previous iteration's M-Step, as in other forms of EM routing, except that in our case votes have two indexes, $ch$, instead of one. See Alg. \ref{alg:Our_EM_Routing_algorithm} for computation. Informally, we can think of $P_{ij}$ as ``the probability of votes from input capsule $i$ given capsule $j$'s probabilistic model.''

In our implementation, for numerical stability, we compute $R_{ij}$ after the first iteration by applying a Softmax function to simplified log-sums, instead of using the second equation in (\ref{eq:equation_for_R}).

\subsection{D-Step}\label{ssec:D_Step}

At the beginning of each D-Step, we multiply the assigned routing probabilities $R_{ij}$ by logistic function activations of input scores, which act as gates, to obtain $D\suptag{use}_{ij}$, the \emph{share of data used} from each input capsule $i$ to update each output capsule $j$'s Gaussian model,

\begin{equation}\label{eq:equation_for_D_use}
D\suptag{use}_{ij} = f \left( a\inp_i \right) R_{ij},
\end{equation}

where $f$ is the logistic function and $a\inp_i$ is the input score associated with input capsule $i$. Except for the ``last-minute'' application of the logistic function, (\ref{eq:equation_for_D_use}) is the same as its corresponding equation in previous forms of EM routing. However, our notation explicitly differentiates between $R_{ij}$, the routing probabilities, and $D\suptag{use}_{ij}$, the share of capsule $i$'s data used by capsule $j$.

Each row of routing probabilities $R_{ij}$ adds up to 1, and $f$ maps $[-\infty, \infty]$ to $[0,1]$; therefore,

\begin{equation}\label{equation_for_D_use_le_ainp_le_1}
0 \le D\suptag{use}_{ij} \le f \left( a\inp_i \right) \le 1,
\end{equation}

that is, $D\suptag{use}_{ij}$ has values that range from 0 (``completely ignore input capsule $i$ in output capsule $j$'s model'') to 1 (``use the whole of input capsule $i$ in output capsule $j$'s model''), but never exceeds $f(a\inp_i)$ (``how much of input capsule $i$ can we use among all output capsules?'').

Given these relationships, we can compute the \emph{share of data ignored} ({\em i.e.}, not used) $D\suptag{ign}_{ij}$ from each input capsule $i$ by each output capsule $j$,

\begin{equation}\label{eq:equation_for_D_ign}
D\suptag{ign}_{ij} = f \left( a\inp_i \right) - D\suptag{use}_{ij},
\end{equation}

such that for each input and output capsule pair $ij$ the two shares, $D\suptag{use}_{ij}$ and $D\suptag{ign}_{ij}$, plus the data that is ``gated off'' by logistic activation of the corresponding input score, $1 - f(a\inp_i)$, add up to 1, or the whole input capsule,

\begin{equation}\label{equation_for_D_use_D_ign_fainp}
D\suptag{use}_{ij} + D\suptag{ign}_{ij} + \left( 1 - f \left( a\inp_i \right) \right) = 1,
\end{equation}

accounting for all input data.

{\em Output capsules are thus in a competition with each other to use ``more valuable bits,'' and ignore ``less valuable bits,'' of input data}. Each output capsule can improve its use-ignore shares only at the expense of other output capsules. 

\subsection{M-Step}\label{ssec:M_Step}

The M-Step computes updated output scores $a\out_j$ and weighted means and variances $\mu\out_{jch}$ and $(\sigma\out_{jch})^2$, respectively, to maximize the probability that each output capsule $j$'s Gaussian model would generate the votes computed from each input capsule $i$ used by $j$. We discuss these computations in the subsections that follow.

\subsubsection{Output Scores}

Previous forms of EM routing use the minimum description length principle to derive approximations of the \emph{cost to activate} and the \emph{cost not to activate} each output capsule $j$, and compute output activations by applying a logistic function to the difference between those approximations. Such costs must be approximated because the only known method for accurately computing them would require inverting $n\inp \times n\out$ vote-computation matrices, which is impractical. See \citet{46653} for details.

We use a different approach, motivated by the intuitive notion that

\begin{quote}
	\emph{``output capsules should benefit from the input data they use, and lose benefits from any input data they ignore,''}
\end{quote}

as they maximize the probability of votes from those input capsules they use in each iteration.

For each output capsule $j$, we compute output score $a\out_j$ as the difference of a \emph{net benefit to use} and a \emph{net cost to ignore} input capsule data,

\begin{equation}\label{eq:equation_for_a}
a\out_j = \showeqfora{\\ \quad}
\end{equation}

where $D\suptag{use}_{ij}$ and $D\suptag{ign}_{ij}$ are the shares of input capsule $i$'s data used and ignored by output capsule $j$, computed in (\ref{eq:equation_for_D_use}) and (\ref{eq:equation_for_D_ign}), respectively, and $\beta\suptag{use}_{ij}$ and $\beta\suptag{ign}_{ij}$ (or $\beta\suptag{use}_j$ and $\beta\suptag{ign}_j$, if $n\inp$ is unspecified) are parameters representing each output capsule's net benefit to use and net cost to ignore input capsule data, respectively. The adjective ``net'' denotes  that $\beta\suptag{use}$ and $\beta\suptag{ign}$ can have positive (``credits'') or negative (``debits'') values.

For certain tasks, we may want the net cost of ignoring each input capsule to be equal to the net benefit we lose from not using it. We can accomplish this trivially by setting $\beta\suptag{ign} = \beta\suptag{use}$ for all $ij$, making them one and the same parameter.

\subsubsection{Output Capsule Probabilistic Models}

At the end of each M-Step, we compute updated means $\mu\out_{jch}$ and variances $(\sigma\out_{jch})^2$ of every output capsule $j$'s Gaussian model, weighted by $D\suptag{use}_{ij}$, the amount of data the output capsule uses from each input capsule $i$. See Alg. \ref{alg:Our_EM_Routing_algorithm} for details.

\subsubsection{Motivation and Intuitition}

In the next iteration's E-Step, when we activate $a\out_j$ by applying to it a logistic function $f$ in (\ref{eq:equation_for_R}), we induce in each output capsule a distribution $(f(a\out_j), 1 - f(a\out_j))$ over a quantity equal to (a) the output capsule's net benefits from those input capsules it uses, less (b) its net costs (or lost benefits) from those input capsules it ignores.

When we recompute routing probabilities $R_{ij}$ in (\ref{eq:equation_for_R}), we weight $P_{ij}$, the probability of votes from input capsules each output capsule uses, by $f(a\out_j)$, {\em the probability of net benefits less costs from using those input capsules}, jointly maximizing both probabilities in the EM loop for optimization of a training objective specified elsewhere.

Informally, we can think of this multi-faceted mechanism as finding, for each output capsule, ``the combination of net benefits and costs that produces greater profit,'' where ``greater profit'' stands for maximizing input vote probabilities at each output capsule and optimizing the whole layer for another objective. We prefer to think of it as maximizing \emph{``bang per bit.''}

\subsubsection{Relationship to Description Length}

There is an interesting connection between our activation mechanism and that used by previous forms of EM routing: All else remaining equal, at each output capsule, using more data from an input capsule is associated with greater description length, and using less data from the input capsule is associated with the opposite---by definition.

\begin{figure*}[t]
	\vskip 0.1in
	\begin{center}
		\centerline{\includegraphics{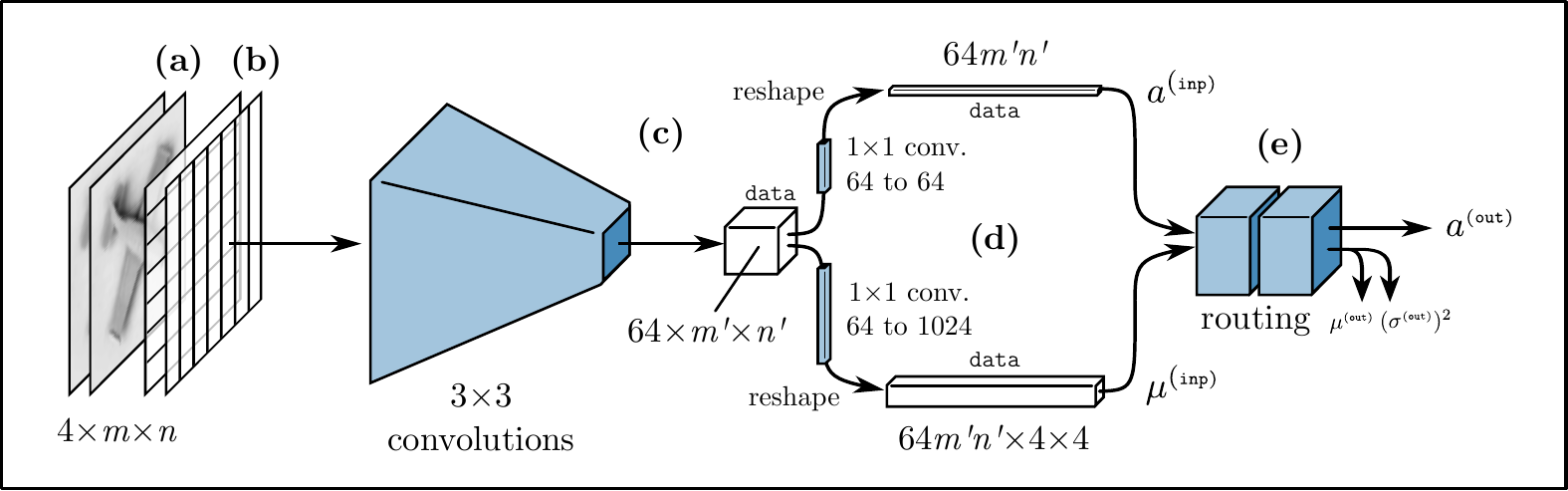}}
		\caption{\textbf{Our smallNORB model}. \textbf{(a)} We stack each pair of images with \textbf{(b)} coordinate values evenly spaced from -1.0 to 1.0, horizontally and vertically, creating an input tensor of shape $4 \times m \times n$, where $m = n = 96$ for unmodified images at test time. \textbf{(c)} We apply six $3 \times 3$ convolutions, each with 64 output channels and alternating strides of 1 and 2. Each convolution is preceded by batch normalization and followed by a Swish activation \cite{DBLP:journals/corr/abs-1710-05941} with constant $\beta = 1$. The last convolution outputs a tensor of shape $64 \times m' \times n'$. \textbf{(d)} We compute $a\inp$ and $\mu\inp$ by applying two $1 \times 1$ convolutions, with 64 and 1024 output channels, respectively, and reshape them as shown. Both convolutions are preceded by batch normalization. After reshaping, $a\inp$ consists of $64m'n'$ input scores, representing possible presence or absence of 64 toy parts in $m'n'$ image locations. $\mu\inp$ consists of $64m'n'$ slices of shape $4 \times 4$, each representing a pose for one of 64 parts in $m'n'$ locations. \textbf{(e)} We apply two layers of our routing algorithm; the first one routes a variable number of input capsules to 64 output capsules, each representing a larger toy part with a $4 \times 4$ pose; the second one routes those capsules to five capsules, each representing a type of toy with a $4 \times 4$ pose. For prediction, we apply a Softmax to $a\out$.}
		\label{fig:smallnorb_model}
	\end{center}
	\vskip -0.2in
\end{figure*} 

\section{Sample Application: smallNORB}
\label{sec:smallNORB_discussion}

The smallNORB dataset \cite{LeCun:2004:LMG:1896300.1896315} has grayscale stereo 96$\times$96 images of five classes of toys: airplanes, cars, trucks, humans, and animals. There are 10 toys in each class, five selected for the training set and five for the test set. Each toy is photographed stereographically at 18 different azimuths (0-340 degrees), 9 different elevations, and 6 lighting conditions, such that the training and test sets each contain 24,300 pairs of images. Supp. Fig. \ref{fig:smallnorb_samples} shows samples from each class.

\subsection{Architecture}

The architecture of our smallNORB model is described in detail in Fig. \ref{fig:smallnorb_model}. At a high level of abstraction, we can think of the model as doing two things: First, it applies a sequence of standard convolutions to detect 64 toy parts and their $4 \times 4$ poses (spatial relationships to the viewer of the image) in multiple possible locations in the image (steps (a) through (d)) in Fig. \ref{fig:smallnorb_model}). Then, the model applies two layers of our routing algorithm, one to detect 64 larger toy parts and their poses, and another to detect five categories of toys and their poses (Fig. \ref{fig:smallnorb_model}(e)). The routing layers are meant to induce the standard convolutions to learn to recognize toy parts and their poses.

\subsubsection{Use of Variable-Size Inputs}

To keep the number of parameters small, we leave $n\inp$ unspecified in the first routing layer, so it accepts a variable number of input capsules without regard for their location in the image. To counteract this loss of location information, we stack input images with two tensors of coordinate values evenly spaced from -1.0 to 1.0, one horizontally and one vertically, as shown in Fig. \ref{fig:smallnorb_model}(b).

Besides reducing the number of parameters in the first routing layer (by a factor of $64m'n'$, as shown in Fig. \ref{fig:smallnorb_model}), our decision to accept a variable number of capsules in that first routing layer makes our model capable of accepting images of variable size, limited only by memory, though we do not make use of this capability here.

\subsection{Initialization and Training}

We initialize all convolutions with Kaiming normal initialization \cite{DBLP:journals/corr/HeZR015} and the two routing layers as follows: Normal initialization scaled by $1 \over d\inp$ for the multilinear weights that compute votes, zeros for the bias parameters, and zeros for the net benefit and cost parameters.

We train the model for 50 epochs, with a batch size of 20, using RAdam \cite{liu2019variance} for optimization via stochastic gradient descent. We use a single-cycle hyperparameter scheduling policy in which learning rate $r$ and first momentum $\beta_1$ start at ($r=10^{-5}$, $\beta_1=0.999$), each change linearly to ($r=5 \times 10^{-4}$, $\beta_1=0.9 \times 0.999$) over the first 10\% of training iterations, and then return to their respective starting values with a cosine shape over the remaining iterations.

During training, we add 16 pixels of padding on each side to each pair of images and randomly crop them to $96 \times 96$ size. We do not use any other image processing, nor any metadata, nor any additional data in training.

For regularization, we use mixup \cite{DBLP:journals/corr/abs-1710-09412} with Beta distribution parameters (0.2, 0.2), inducing the iterative EM clustering algorithms in our routing layers to learn to distinguish samples that have been mixed together.

The objective function is a Cross Entropy loss, computed on Softmax activations of the output scores of the final routing layer's five capsules. Supp. Fig. \ref{fig:validation_plots} shows validation loss and accuracy after each epoch of training.

\subsection{Results}

Tab. \ref{table:smallnorb} shows test set accuracy, number of parameters, and number of epochs to train our smallNORB model, and how it compares to prior models that have achieved state-of-the-art results without using any metadata or additional data.

\begin{table}[h]
	\small
	\begin{center}
		\begin{tabular}{@{}lccc@{}}
			\toprule
			\bf &  \bf Accuracy &\bf No. of & \bf Train\\
			\bf Model & \bf (\%) & \bf Params & \bf Epochs \\
			\midrule
			\citet{DBLP:journals/corr/abs-1710-09829} & 97.3 & 8200K & N/A \\
			\citet{DBLP:journals/corr/abs-1102-0183}* & 97.4 & 2700K & N/A \\
			\citet{46653} & 98.2 & 310K & 300 \\
			\citet{46653}** & 98.6 & 310K & 300 \\
			Our Model & \bf 99.1 & \bf 272K & \bf 50 \\
			\bottomrule
		\end{tabular}
	\end{center}
	\begin{tabular}{l}
		\scriptsize * Extensive image processing in training. \\
		\scriptsize ** Reported accuracy is mean of multiple random crops per image. \\
	\end{tabular}
	\caption{\label{table:smallnorb}Test set accuracy, parameters, and number of training epochs of models with state-of-the-art performance on smallNORB.}
\end{table}

Compared to the previous state of the art \cite{46653}, our model has fewer parameters, trains in an order of magnitude fewer epochs, and accepts full-size unprocessed images instead of downsampled, cropped ones. We do not use multiple crops per image to compute test accuracy.

Compared to the best-performing conventional CNN on this benchmark \cite{DBLP:journals/corr/abs-1102-0183}, our model has an order of magnitude fewer parameters and is trained with minimal data augmentation (only cropping), whereas the CNN is trained with additional stereo pairs of images created using different filters and affine distortions.

\subsection{Analysis}

\begin{figure}[h]
	\vskip 0.1in
	\begin{center}
		\centerline{\includegraphics{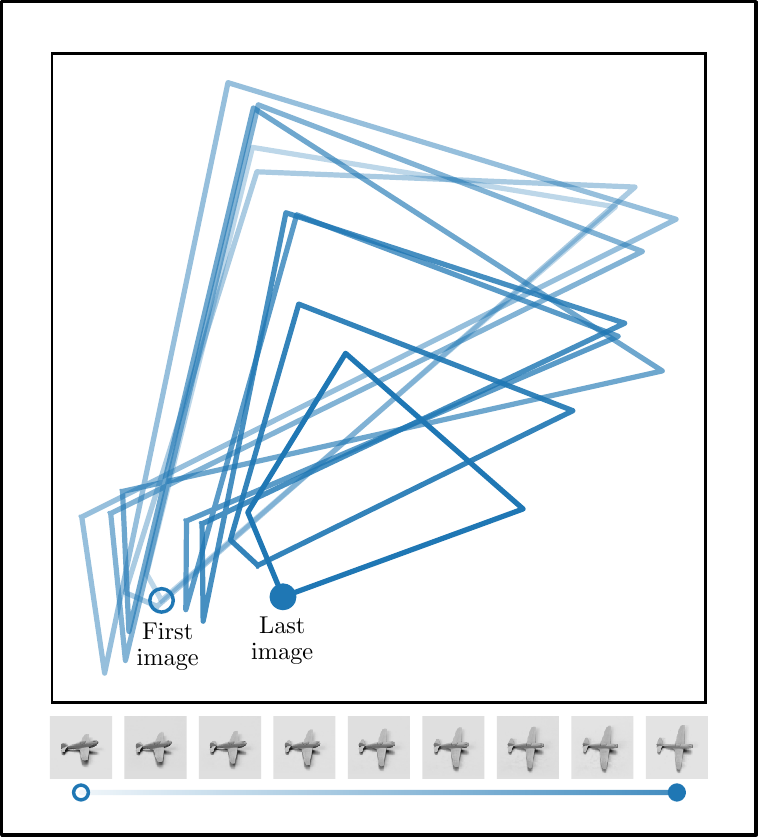}}
		\caption{Multidimensional scaling (MDS) representations in $\mathbb{R}^2$ of the trajectories of an activated class capsule's four pose vectors, each of size $d\out = 4$, as we feed test images of an object in the class with varying elevations to our trained smallNORB model. For each image, the $\mathbb{R}^2$ coordinates are plotted as four connected vertices, each vertex corresponding to a pose vector, preserving as much as possible the pairwise distances between pose vectors from all images. Circles indicate the activated capsule's first pose vector for the first and last image.}
		\label{fig:smallnorb_elevation_trajectories}
	\end{center}
	\vskip -0.2in
\end{figure}

Besides superior test set performance, we find evidence that our smallNORB model learns to perform its own form of ``reverse graphics'' without explicitly optimizing for it, solely from pixel data and classification labels. The model learns to use all four pose vectors jointly to represent poses, albeit in a way that feels quite alien compared to the typical human approach ({\em e.g.}, a $3 \times 3$ rotation matrix inside a $4 \times 4$ matrix with translation data).

The visualization in Fig. \ref{fig:smallnorb_elevation_trajectories} shows multidimensional scaling (MDS) representations in $\mathbb{R}^2$ of the trajectories of an activated class capsule's pose vectors as we feed test images of an object in the class with varying elevations to our model. We can see that the four pose vectors that constitute the class capsule jointly move and eventually seem to ``flip'' as we change viewpoint elevation. The same visualization for other objects in the dataset, and for varying azimuth, look qualitatively similar.

We also analyze quantitatively the behavior of pose vectors as we vary azimuth and elevation for every category and instance of toy in the dataset, and find that pose vectors behave in ways that are consistent with variation in azimuth and elevation. See the 24 plots and their captions in Supp. Fig. \ref{fig:smallnorb_change_in_pose_vectors_azimuth} and Supp. Fig. \ref{fig:smallnorb_change_in_pose_vectors_elevation} for details.

Much more work remains to be done to understand and quantify our routing algorithm's ability to learn ``reverse graphics.'' However, we think such work falls outside the scope of this paper, given that we also evaluate our routing algorithm in another domain, natural language.

\begin{figure*}[t]
	\vskip 0.1in
	\begin{center}
		\centerline{\includegraphics{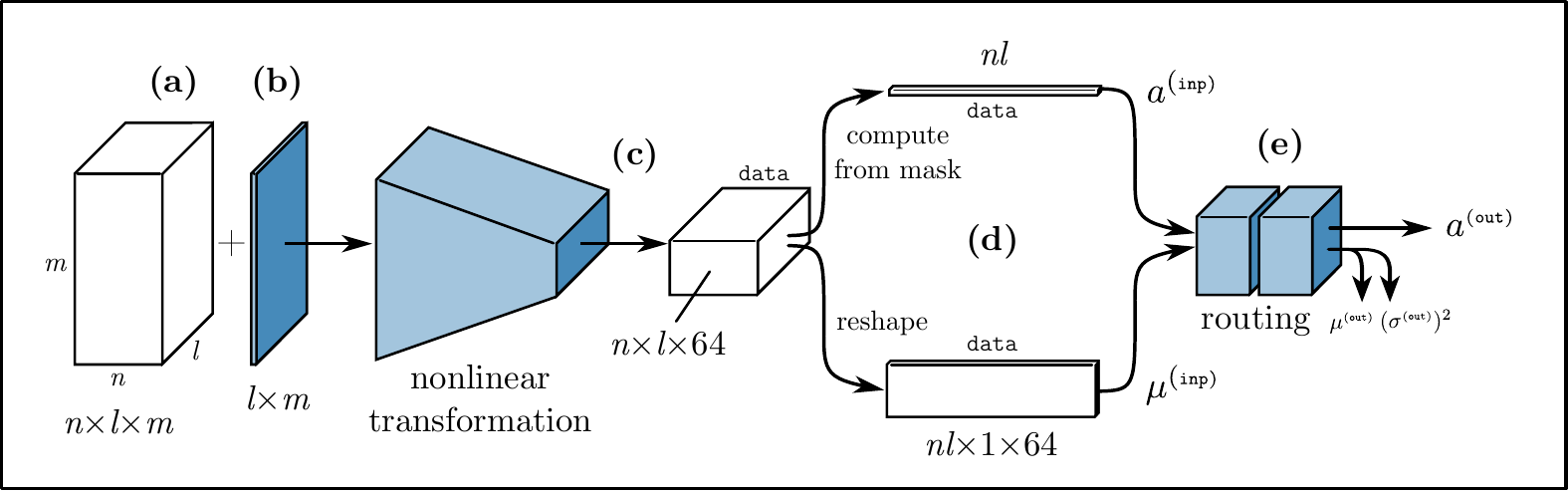}}
		\caption{\textbf{Our SST model.} \textbf{(a)} For each sample, the input is a tensor of transformer embeddings of shape $n \times l \times m$, where $n$ is the number of tokens, $l$ is the number of transformer layers, and $m$ is the embedding size. \textbf{(b)} We element-wise add to the input tensor a depth-of-layer parameter of shape $l \times m$. \textbf{(c)} We apply a linear transformation from $m$ to 64, followed by a Swish activation with constant $\beta = 1$ and layer normalization, obtaining a tensor of shape $n \times l \times 64$. \textbf{(d)} We reshape the tensor as shown to obtain $\mu\inp$, consisting of $ln$ input capsules of size $1 \times 64$. We compute $a\inp \longleftarrow \log(\frac{x}{1 -x})$ from a mask $x$ of length $nl$ with ones and zeros indicating, respectively, which embeddings correspond to tokens and which correspond to any padding necessary to group samples in batches, obtaining logits that are equal to $\infty$ for tokens, $-\infty$ for padding, and values in between for any tokens and padding that get combined by mixup regularization in training. \textbf{(e)} We apply two layers of our routing algorithm; the first one routes a variable number of capsules in  $\mu\inp$ to 64 capsules of shape $1 \times 2$; the second one routes those capsules to five or two capsules of equal shape, each representing a classification label in SST-5/R or SST-2/R. For prediction, we apply a Softmax to output scores $a\out$.}
		\label{fig:SST5R_model}
	\end{center}
	\vskip -0.2in
\end{figure*}

\section{Sample Application: SST}
\label{sec:SST5R_discussion}

The Stanford Sentiment Treebank (SST) \cite{brusilovsky:socher2013recursive} consists of 11,855 sentences extracted from movie reviews, parsed into trees with 215,154 unique phrases. The root sentences are split into training (8,544), validation (1,101) and test (2,210) sets, each with its own subset of the parse trees. The fine-grained root sentence classification task (SST-5/R) involves selecting one of five labels ({\tt \small very negative, negative, neutral, positive, very positive}) for each root sentence in the test set. The binary root sentence classification task (SST-2/R) involves selecting one of two labels ({\tt \small negative, positive}) after removing all {\tt \small neutral} samples from the dataset, leaving 9,613 root sentences, split into training (6,920), validation (872), and test (1,821) sets, each with their own subset of the parse trees.

We chose SST, and SST-5/R in particular, for three reasons: First, its size is small enough to sidestep certain challenges to scaling EM routing ({\em e.g.}, see \citet{Barham:2019:MLS:3317550.3321441}). Second, since this dataset's release in 2013, no model has come close to human performance on SST-5/R, as measured by accuracy on its labels, which were assigned by human beings. Finally, we suspect SST-5/R has remained challenging because it is less susceptible than other benchmarks to the ``Clever Hans'' effect, in which seemingly impressive performance is explained by exploitation of spurious statistical cues in the data.

The Clever Hans effect has been documented in multiple natural language datasets, for example, by \citet{mccoy-etal-2019-right} and \citet{DBLP:journals/corr/abs-1907-07355}. Models have become so good at recognizing patterns in natural language that human beings are now finding it difficult to design benchmarks that are free of spurious statistical cues.

Several qualities, we think, make it challenging for machines to find and exploit spurious statistical cues in SST-5/R: First, its labels map to sentiments that transition into each other in complicated ways ({\em e.g.}, the boundary between {\tt \small neutral} and {\tt \small positive} sentences). Second, the dataset is small ({\em e.g.}, only 2,210 test sentences). Third, the samples exhibit a variety of complex syntactic phenomena ({\em e.g.}, nested negations). Finally, the samples exhibit diverse linguistic constructions ({\em e.g.}, idiosyncratic movie fan idioms).

\subsection{Architecture}

The architecture of our SST model resembles that of our smallNORB model. We show and describe it in detail in Fig. \ref{fig:SST5R_model}. At a high level of abstraction, we can think of our SST model as doing two things: First, it applies a nonlinear transformation to every embedding from a pretrained transformer, mapping each one to a vector with 64 elements indicating present or absence of ``sentiment features'' (steps (a) through (d)) in Fig. \ref{fig:SST5R_model}). Then, the model applies two layers of our routing algorithm, one to detect 64 composite parts, and another to detect classification labels, five for SST-5/R and two for SST-2/R (Fig. \ref{fig:SST5R_model}(e)). The routing layers are meant to induce the nonlinear transformation to learn to recognize useful ``sentiment parts.''

\subsubsection{Use of Variable-Size Inputs}

The number of tokens in sentences is variable, so we leave $n\inp$ unspecified in the first routing layer of our SST model. This first routing layer accepts any number of input capsules without regard for their position in the sequence or the depth of the transformer layer from which they originate.

Transformer embeddings incorporate information about their position in a sequence, but not about layer depth. To counteract the loss of depth information, we add a ``depth-of-layer'' parameter to the input tensor, as shown in Fig. \ref{fig:SST5R_model}(b). In this parameter, each transformer layer has a corresponding vector slice, which we add element-wise to every embedding in the input tensor originating from that transformer layer.

\subsubsection{Use of GPT-2}

In our implementation, we use a GPT-2$_\text{large}$ \cite{GPT-2-Radford-et-al} as the pretrained transformer. We chose this model mainly because it was the largest one publicly available at the time of writing, and also because we like the simplicity of its training objective: it is trained only to predict the next subword token in approximately 40GB of text.

GPT-2$_\text{large}$ has approximately 774 million parameters and outputs 37 layers (36 hidden plus one visible) of dimension 1280. If a sentence contains, say, 10 tokens, this GPT-2 model transforms it into $37$ sequences, each with 10 embeddings of size 1280. We concatenate them into an input tensor of shape $10 \times 37 \times 1280$ for our SST model.

\subsection{Initialization and Training}

We initialize the routing layers {\em in the same manner} as for the smallNORB model, the depth embeddings with zeros, and the linear transformations with Kaiming normal.

The training regime is {\em the same} as for the smallNORB model, except that we train the SST model for only 3 epochs. We use as training data all unique token sequences in the parse trees of the training split.  Supp. Fig. \ref{fig:validation_plots} shows loss and accuracy on the root-sentence validation set after each epoch of training for both SST-5/R and SST-2/R.

\subsection{Results on SST-5/R}

As Tab. \ref{tab:SST5R} shows, we achieve new state-of-the-art test set accuracy of 58.5\% on SST-5/R, a significant improvement (2.3 percentage points) over previous state-of-the-art test set accuracy.

\begin{table}[h]
	\small
	\begin{center}
		\begin{tabular}{@{}lc@{}}
			\toprule
			\bf Model/Pretraining & \bf SST-5/R (\%) \\
			\midrule
			RNTN/{\scriptsize none} (Socher et al., 2013) & 45.7 \\
			CNN/{\scriptsize word2vec} (Kim et al., 2014) & 48.0 \\
			Para-Vec/{\scriptsize on dataset} (Le and Mikolov, 2014) & 48.7 \\
			LSTM/{\scriptsize on PP2B} (Wieting et al., 2015) & 49.1 \\
			DMN/{\scriptsize GloVe} (Kumar et al., 2015) & 51.1 \\
			NSE/{\scriptsize GloVe} (Mundkhdalai and Yu, 2016) & 52.8 \\
			ByteLSTM/{\scriptsize 82M reviews} (Radford et al., 2017) & 52.9 \\
			CT-LSTM/{\scriptsize word2vec} (Looks et al., 2017) & 53.6 \\
			BCN+Char/{\scriptsize CoVe} (McCann et al., 2017) & 53.7 \\
			BCN/{\scriptsize ELMo} (Peters et al., 2018) & 54.7 \\
			SuBiLSTM+Char/{\scriptsize CoVe} (Brahma, 2018) & 56.2 \\
			Our Model/{\scriptsize GPT-2$_\text{large}$ (non-finetuned)} & \bf 58.5 \\
			\bottomrule
		\end{tabular}
	\end{center}
	\caption{\label{tab:SST5R}State-of-the-art test set accuracy on SST-5/R since its publication is associated with both new architectures and new pretraining methods.}
\end{table}

\subsection{Results on SST-2/R}

Changing only the final number of capsules to two, the same model achieves test set accuracy of 95.6\% on SST-2/R, a new state-of-the-art performance for single-task models. The previous state of the art was achieved by BERT \cite{DBLP:journals/corr/abs-1810-04805}. Current state-of-the-art test set accuracy for multi-task models or ensembles is 96.8\%, by an ensemble of XLNet models trained on multiple GLUE tasks \cite{DBLP:journals/corr/abs-1906-08237}. See Tab. \ref{tab:SST2R}.

\begin{table}[h]
	\small
	\begin{center}
		\begin{tabular}{@{}lc@{}}
			\toprule
			\bf Model & \bf SST-2/R (\%) \\
			\midrule
			\em Multi-task models or ensembles: &  \\
			Snorkel (Ratner et al., 2018) (ensemble) & 96.2 \\
			MT-DNN (Liu et al., 2019) (single model) & 95.6 \\
			MT-DNN (Liu et al., 2019) (ensemble) & 96.5 \\
			XLNet (Yang et al., 2019) (ensemble) & \bf 96.8 \\
			\midrule
			\em Single-task models: & \\
			BCN+Char/{\scriptsize CoVe} (McCann et al., 2017) & 90.3 \\		
			Block-sparse LSTM (Radford et al., 2017) & 93.2 \\
			BERT (Devlin et al., 2018) & 94.9 \\
			Our Model/{\scriptsize GPT-2$_\text{large}$ (non-finetuned)} & \bf 95.6 \\
			\bottomrule
		\end{tabular}
	\end{center}
	\caption{\label{tab:SST2R}Recent state-of-the-art test set accuracy on SST-2/R by multi-task models or ensembles, and by single-task models like ours.}
\end{table}

\subsection{Analysis}

We are surprised to see a such a large improvement over the state of the art on SST-5/R and also new state-of-the-art performance on SST-2/R compared to previous single-task models, considering that (a) we do {\em not} finetune the transformer, (b) our SST model resembles the model we use for a visual task, and (c) we train the SST model with the same regime, changing only the number of epochs. (By implication, greater accuracy might be possible with transformer finetuning and more careful tweaking of model and training regime.)

We also note that progress on SST-5/R has been remarkably steady, year after year, since its publication in 2013, as AI researchers have devised new architectures that exploit new pretraining mechanisms (Tab. \ref{tab:SST5R}). Our results represent a continuation of this long-term trend. As of yet, no model has come close to accurately modeling the labeling decisions of human beings on SST-5/R. Performance on the binary task, SST-2/R, whose labels lack the subtlety of the fine-grained ones, has been close to human baseline for several years now.

Finally, our successful use of a capsule network to route embeddings from a pretrained transformer links two areas of AI research that have been largely independent from each other: capsule networks with routing by agreement, used mainly for visual tasks, and transformers with self-attention, used mainly for sequence tasks.

\section{Related Work}\label{related_work}

\citet{46653} proposed the first form of EM routing and showed that capsule networks using it to route matrix capsules can generalize to different poses of objects in images and resist white-box adversarial attacks better than conventional CNNs. Their ``related work'' section compares capsule networks to other efforts for improving the ability of visual recognition models to deal effectively with viewpoint variations.

\citet{DBLP:journals/corr/abs-1710-09829} showed that capsule networks with an earlier form of routing by agreement, operating on vector capsules, can be more effective than conventional CNNs for segmenting highly overlapping images of digits.

\citet{Barham:2019:MLS:3317550.3321441} showed that currently it can be challenging to scale capsule networks to large datasets and output spaces in some circumstances due to current software ({\em e.g.}, PyTorch, TensorFlow) and hardware ({\em e.g.}, GPUs, TPUs) systems, which are highly optimized for a fairly small set of computational kernels, in a way that is tightly coupled with memory hardware, leading to poor performance on non-standard workloads, including basic operations on capsules.

\citet{DBLP:journals/corr/abs-1906-02715} found evidence that BERT, and possibly other transformer architectures, learn to embed sequences of natural language as \emph{trees}. Their work inspired us to wonder if capsule networks might be able to recognize such ``language trees'' in different ``poses,'' analogously to the way in which capsule networks can recognize different poses of objects embedded in images.

\citet{DBLP:journals/corr/VaswaniSPUJGKP17} proposed transformer models using query-key-value dot-product attention, and showed that such models can be more effective than prior methods for modeling sequences. Our routing algorithm can be seen as a new kind of attention mechanism in which output capsules ``compete with each other for the attention of input capsules,'' with each output capsule seeing a different set of input capsule votes.

\section{Future Work}

Capsule networks are a recent innovation, and our routing algorithm is still more recent. Its behavior and properties are not yet widely or fully understood. As current challenges to scaling, such as those studied by \citet{Barham:2019:MLS:3317550.3321441}, are overcome, we think it would make sense to conduct more comprehensive evaluations and ablation studies of our algorithm in multiple domains.

We are also intrigued about using our routing algorithm for natural language modeling. At present this seems impractical, due in part to the computational complexity of the algorithm.\footnote{
	Consider: if we wanted to use our routing algorithm to predict the next capsule in a natural language sequence, over a dictionary of typical size, say, $3 \times 10^4$ subword ids, the final layer of the model alone would have to compute and hold in memory the equivalent of $3 \times 10^4$ simultaneous EM loops, each on a different set of input votes per output capsule.
} A more tractable alternative in the near future might be to intersperse layers of our routing algorithm between blocks of query-key-value self-attention.

Another possible avenue for future research involves experimenting with probabilistic models other than a multidimensional Gaussian in output capsules. While our limited experiments show that a multidimensional Gaussian works remarkably well, we harbor some doubts about its effectiveness with capsules of much greater size.

Finally, we naturally wonder about using non-probabilistic clustering in our routing algorithm, k-means being the most obvious choice, given its relationship to EM and its proven effectiveness at dealing with large-scale data in other settings.

\section{Conclusion}

Building on recent work on capsule networks, we propose a new, general-purpose form of routing by agreement that computes output capsule activations as a logistic function of net benefits to use less net costs to ignore input capsules. To make the computation of net benefits and costs possible, we introduce a new step in the EM loop, the D-Step, that computes the share of data used and ignored from each input capsule by each output capsule, accounting for all input capsule data. We construct our routing algorithm to accept variable-size inputs, such as sequences, which also proves useful for keeping the number of model parameters small in applications for which it is otherwise not necessary. We also explain how to adapt the algorithm for variable-size outputs. Finally, our algorithm uses ``pre-activation'' scores to which we apply logistic functions as needed, facilitating more flexible use by subsequent layers and/or objective functions, with more numerical stability.

We illustrate the usefulness of our routing algorithm with two capsule networks that apply it in different domains, vision and language. Both networks achieve state-of-the-art performance in their respective domains after training with the same regime, thereby showing that {\em adding one or more layers of our routing algorithm can produce state-of-the-art results in more than one domain, without requiring tuning}. Our motivation is to develop universal, composable learning algorithms. Our work is but a small step in this direction.

\section*{Acknowledgments}

We thank Russell T. Klophaus for his feedback.

\bibliography{An_Algorithm_for_Routing_Capsules_in_All_Domains}
\bibliographystyle{An_Algorithm_for_Routing_Capsules_in_All_Domains}

\appendix

\section*{List of Supplementary Figures}

\begin{figure*}[h]
	\begin{center}
		\centerline{\includegraphics{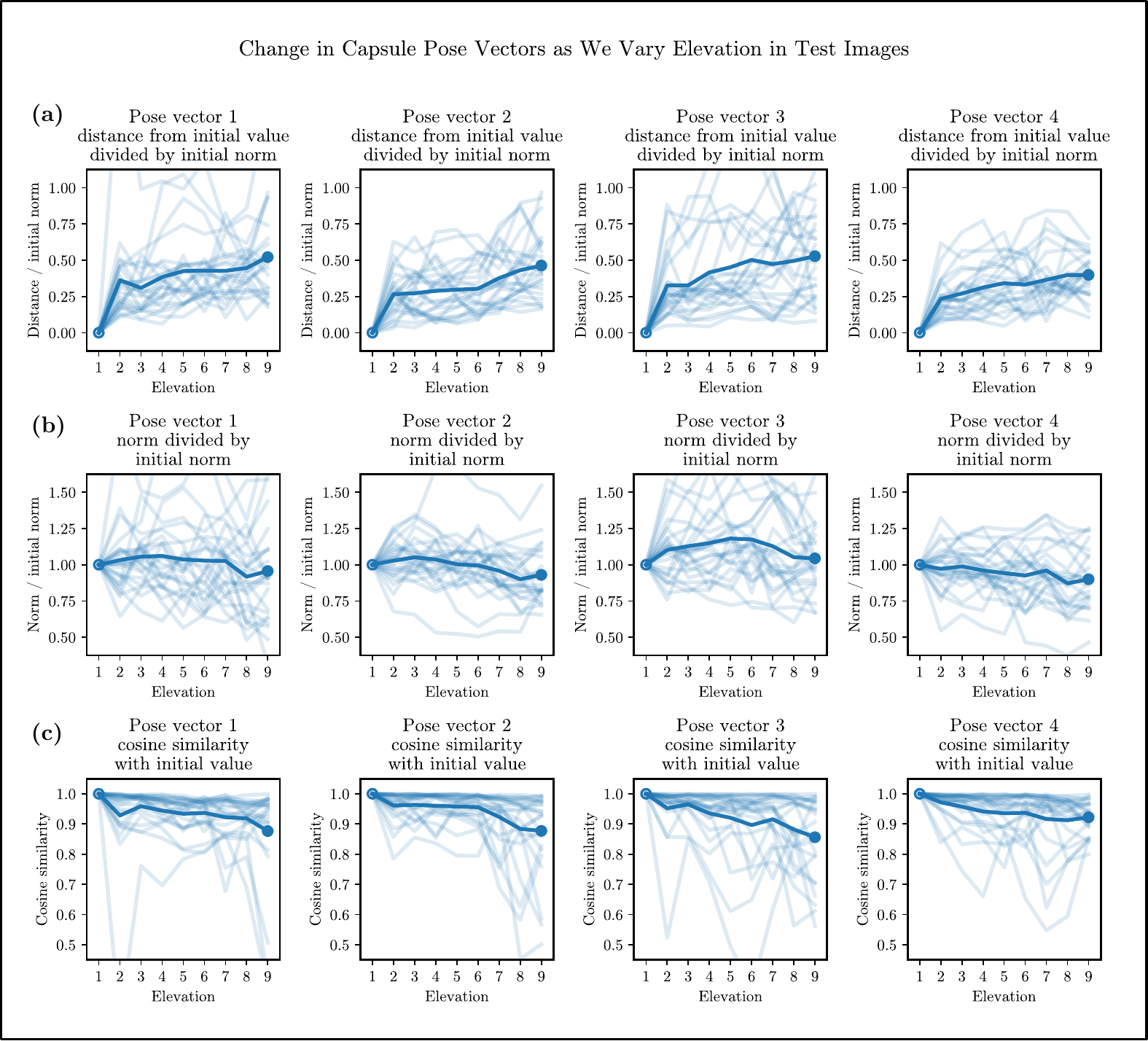}}
		\caption{For each toy category and instance in the test set, we feed our smallNORB model a sequence of images with varying elevation, while keeping everything else constant, and plot the change in each pose vector of the activated capsule. The mean change is shown in dark blue. {\bf (a)} The first row of plots show each pose vector's Euclidean distance to its initial value, divided by the norm of the initial value, as elevation varies through nine levels (from near flat to looking from above). We can see that the pose vector moves away from its initial value, consistent with the change in elevation. {\bf (b)} The second row shows the norm of all pose vectors, divided by their initial norms. We can see that pose vector norms tend to stay close to the initial norm, consistent with rotation due to the change in elevation. {\bf (c)} The third row shows cosine similarity between each pose vector and its initial value. We can see that the angle tends to increase but not decrease back to its original value, consistent with the change in elevation.}
		\label{fig:smallnorb_change_in_pose_vectors_elevation}
		\end{center}
	\end{figure*}

\begin{figure*}[h]
	\begin{center}
		\centerline{\includegraphics{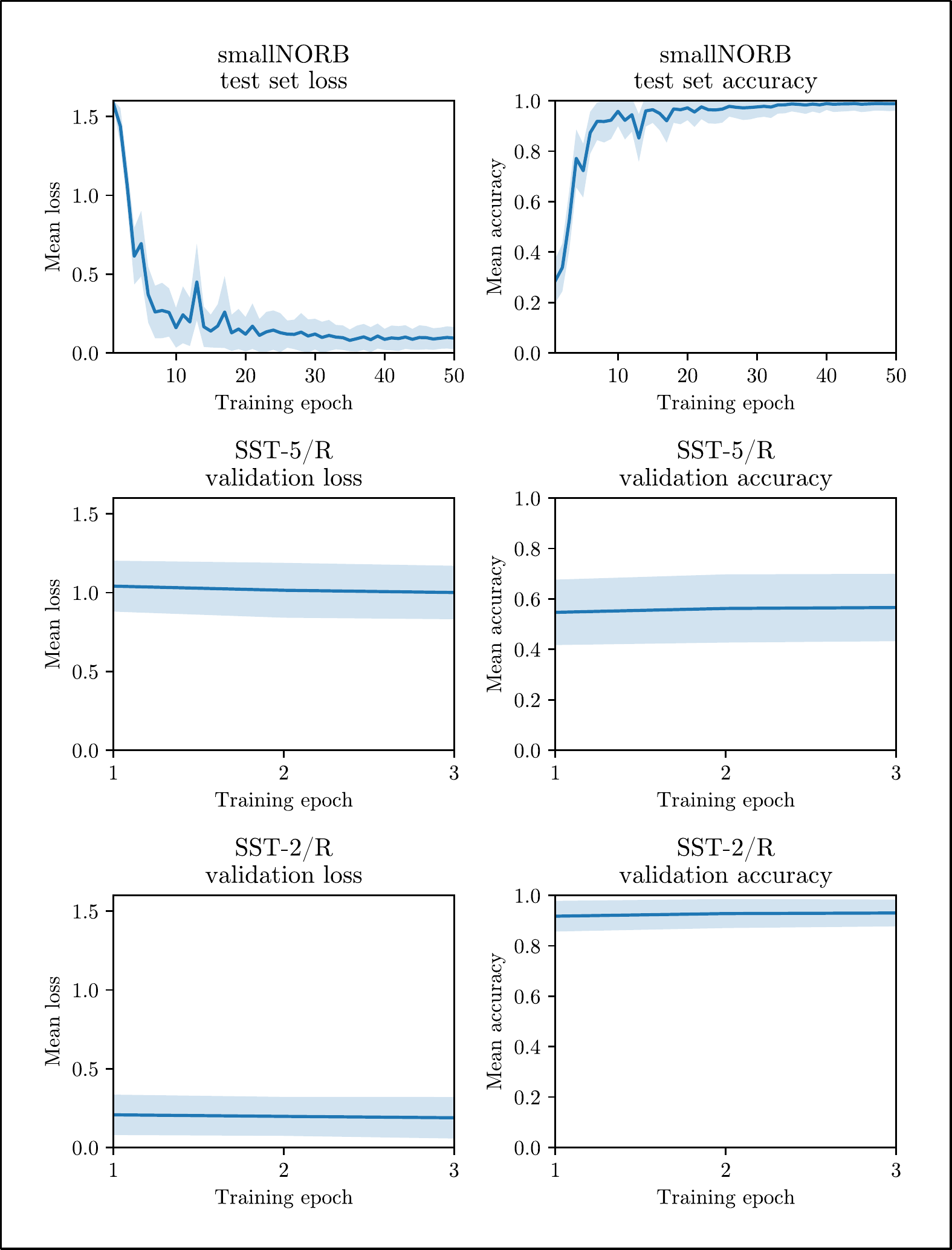}}
		\caption{Mean validation loss and accuracy after each epoch of training. Shaded area denotes standard deviation of batches, with 20 samples each. Note: smallNORB dataset does not have a validation split.}
		\label{fig:validation_plots}
	\end{center}
\end{figure*}

\begin{figure*}[h]
	\begin{center}
		\centerline{\includegraphics{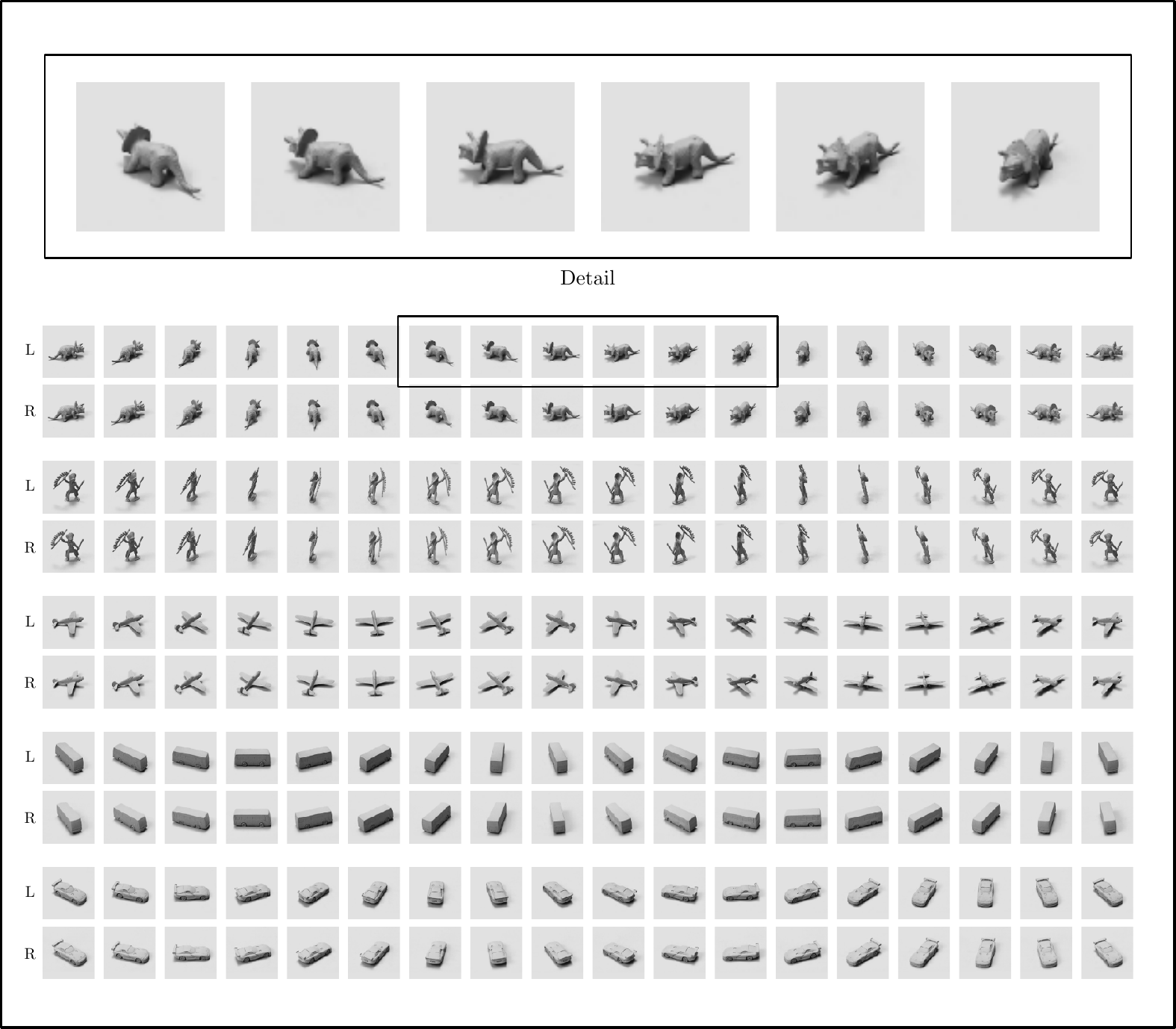}}
		\caption{Sample smallNORB stereographic images of one toy in each of five toy categories. For each toy we show 18 image-pair samples of varying azimuth while keeping elevation and lighting constant.}
		\label{fig:smallnorb_samples}
	\end{center}
\end{figure*}

\begin{itemize}
	\item Fig. \ref{fig:smallnorb_change_in_pose_vectors_azimuth}: Change in pose vectors by azimuth.
	\item Fig. \ref{fig:smallnorb_change_in_pose_vectors_elevation}: Change in pose vectors by elevation.
	\item Fig. \ref{fig:validation_plots}: Validation losses and accuracies.
	\item Fig. \ref{fig:smallnorb_samples}: Sample smallNORB images.
\end{itemize}

\begin{figure*}[h]
	\begin{center}
		\centerline{\includegraphics{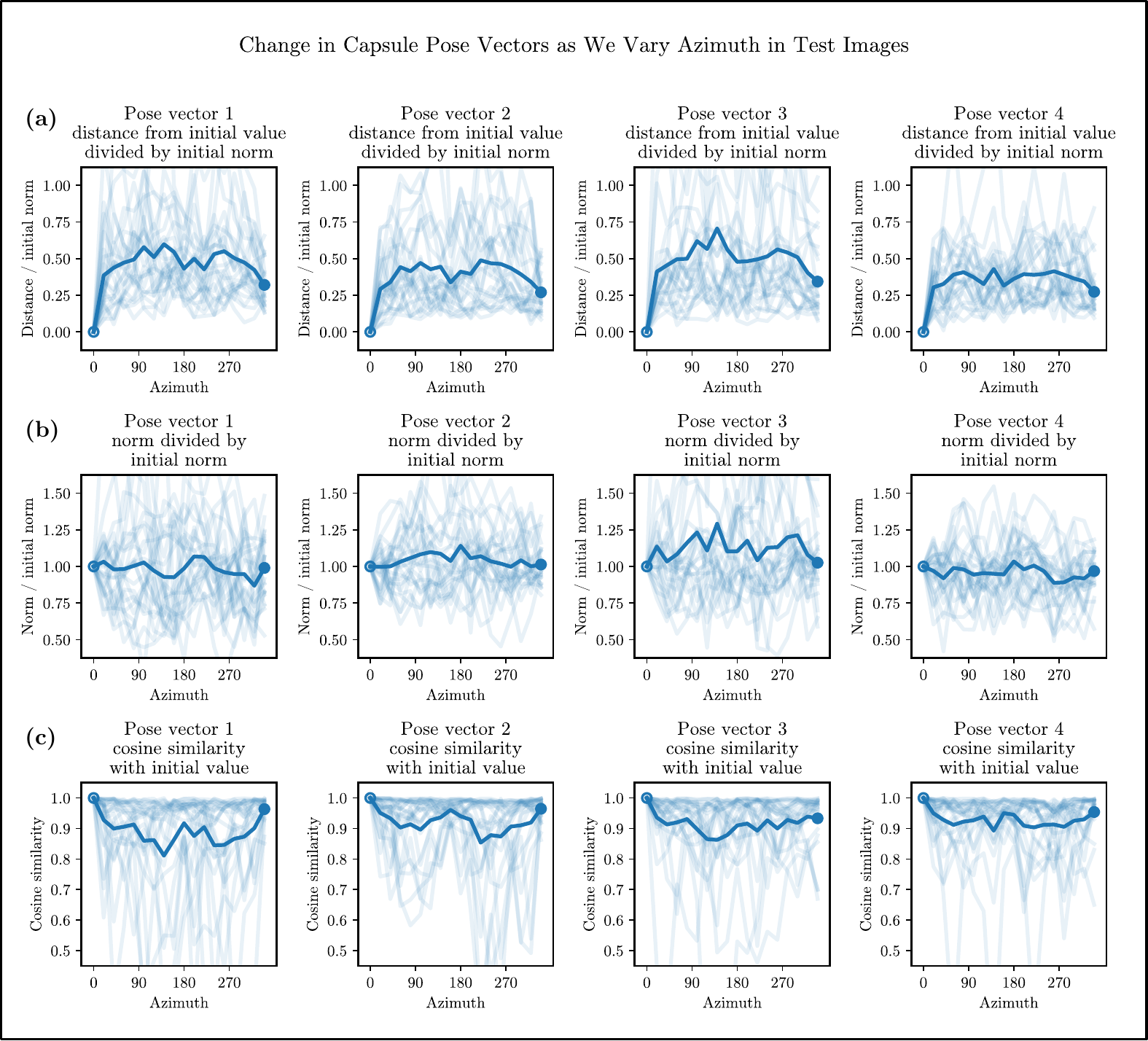}}
		\caption{For each toy category and instance in the test set, we feed our smallNORB model a sequence of images with varying azimuth, while keeping everything else constant, and plot the change in each pose vector of the activated capsule. The mean change is shown in dark blue. {\bf (a)} The first row of plots show each pose vector's Euclidean distance to its initial value, divided by the norm of the initial value, as azimuth varies from 0 to 340 degrees. We can see that the pose vector tends to move away from and then back close to its initial value, consistent with rotation. {\bf (b)} The second row shows the norm of all pose vectors, divided by their initial norms. We can see that pose vector norms tend to stay close to the initial norm, consistent with rotation. {\bf (c)} The third row shows cosine similarity between pose vectors and their initial value. We can see that the angle tends to increase and then decrease, consistent with rotation.}
		\label{fig:smallnorb_change_in_pose_vectors_azimuth}
	\end{center}
\end{figure*}

\end{document}